\documentclass[10pt]{article}
\usepackage{newpxtext}
\usepackage{ascmac}
\usepackage{amsmath}
\usepackage{amssymb}
\usepackage{amsthm}
\usepackage{graphicx}
\usepackage{latexsym}
\usepackage{color}
\usepackage{amssymb}
\usepackage{bm}
\usepackage{subfigure}
\usepackage[left = 20 truemm , right = 20 truemm]{geometry}
\usepackage{url}
\usepackage{algorithmic}
\usepackage{algorithm}
\usepackage{dcolumn}
\usepackage{physics}
\usepackage{caption}
\usepackage{orcidlink}

\usepackage[backend=biber, sorting=none, maxnames=99, firstinits=true, citestyle=numeric, giveninits]{biblatex}
\DeclareFieldFormat[article,book,inbook,incollection,inproceedings,patent,thesis,unpublished]{title}{#1}
\renewbibmacro{in:}{}
\DeclareNameAlias{author}{last-first}

\renewrobustcmd*{\bibinitperiod}{}
\renewrobustcmd*{\bibinitdelim}{}
\renewrobustcmd*{\bibinithyphendelim}{}
\DeclareFieldFormat{pages}{: #1}
\DeclareFieldFormat{volume}{#1}
\DeclareFieldFormat[article]{number}{\mkbibparens{#1}}

\DeclareFieldFormat{volpages}{%
  \printfield{volume}%
  \iffieldsequal{number}{}
    {}
    {
      \setunit{\addspace}%
      \printfield{number}
    }
}

\DeclareBibliographyDriver{inproceedings}{%
  \printnames{author}%
  \setunit{\addspace}%
  \printtext{(\printfield{year})}%
  \newunit\newblock
  \printfield{title}%
  \newunit\newblock
  \printfield{booktitle}%
  \setunit{\addcomma\space}%
  \printfield[volpages]{volume}
  \printfield{pages}%
  \setunit{\addcomma\space}%
  \printlist{publisher}%
  \finentry
}

\DeclareBibliographyDriver{inbook}{%
  \printnames{author}%
  \setunit{\addspace}%
  \printtext{(\printfield{year})}%
  \newunit\newblock
  \printfield{title}%
  \newunit\newblock
  \printfield{booktitle}%
  \setunit{\addcomma\space}%
  \printfield[volpages]{volume}
  \printfield{pages}%
  \setunit{\addcomma\space}%
  \printlist{publisher}%
  \setunit{\addcomma\space}%
  \printfield{doi}%
  \finentry
}

\DeclareBibliographyDriver{misc}{%
  \printnames{author}%
  \setunit{\addspace}%
  \printtext{(\printfield{year})}%
  \newunit\newblock
  \printfield{title}%
  \newunit\newblock
  \printfield{howpublished}%
  \finentry
}

\DeclareBibliographyDriver{article}{%
  \printnames{author}%
  \setunit{\addspace}%
  \printtext{(\printfield{year})}%
  \newunit\newblock
  \printfield{title}%
  \newunit\newblock
  \printfield{journaltitle}%
  \setunit{\addcomma\space}%
  \printfield[volpages]{volume}
  \unspace
  \iffieldsequal{pages}{}
    {}
    {
      \printfield{pages}
    }
  \setunit{\addcomma\space}%
  \printfield{doi}%
  \finentry
}

\usepackage{doi}
\usepackage[T1]{fontenc}
\newenvironment{affiliations}{%
    \setcounter{enumi}{1}%
    \setlength{\parindent}{0in}%
    \slshape\sloppy%
    \begin{list}{\upshape$^{\arabic{enumi}}$}{%
        \usecounter{enumi}%
        \setlength{\leftmargin}{0in}%
        \setlength{\topsep}{0in}%
        \setlength{\labelsep}{0in}%
        \setlength{\labelwidth}{0in}%
        \setlength{\listparindent}{0in}%
        \setlength{\itemsep}{0ex}%
        \setlength{\parsep}{0in}%
        }
    }{\end{list}\par\vspace{12pt}}

\title{Asynchronous Batch Bayesian Optimization with Pipelining Evaluations for Experimental Resource\textendash constrained Conditions}
\author{Yujin Taguchi$^{1}$ \orcidlink{0009-0007-7997-6880},
Yusuke Shibuya$^{1}$,
Yusuke Hiki$^{1}$ \orcidlink{0000-0001-6955-3867}, \\
Takashi Morikura$^{2}$ \orcidlink{0000-0003-1287-8809},
Takahiro G. Yamada$^{1,2}$ \orcidlink{0000-0002-1665-1778},
Akira Funahashi$^{1,2,*}$ \orcidlink{0000-0003-0605-239X}}
\date{}

\begin{document}
\maketitle

\begin{affiliations}
 \item[1] \,Center for Biosciences and Informatics, Graduate School of Fundamental Science and Technology, Keio University, Kanagawa, Japan
 \item[2] \,Department of Biosciences and Informatics, Keio University, Kanagawa, Japan
 \item[*] \,Corresponding author: A. F. (funa@bio.keio.ac.jp)
\end{affiliations}

\section*{Abstract}

Bayesian optimization is efficient even with a small amount of data and is used in engineering and in science, including biology and chemistry.
In Bayesian optimization, a parameterized model with an uncertainty is fitted to explain the experimental data, and then the model suggests parameters that would most likely improve the results.
Batch Bayesian optimization reduces the processing time of optimization by parallelizing experiments.
However, batch Bayesian optimization cannot be applied if the number of parallelized experiments is limited by the cost or scarcity of equipment;
 in such cases, sequential methods require an unrealistic amount of time. In this study, we developed pipelining Bayesian optimization (PipeBO) to reduce the processing time of optimization even with a limited number of parallel experiments.
PipeBO was inspired by the pipelining of central processing unit architecture, which divides computational tasks into multiple processes.
PipeBO was designed to achieve experiment parallelization by overlapping various processes of the experiments.
PipeBO uses the results of completed experiments to update the parameters of running parallelized experiments.
Using the Black-Box Optimization Benchmarking, which consists of 24 benchmark functions, we compared PipeBO with the sequential Bayesian optimization methods.
PipeBO reduced the average processing time of optimization to about 56\% for the experiments that consisted of two processes or even less for those with more processes for 20 out of the 24 functions.
Overall, PipeBO parallelizes Bayesian optimization in the resource-constrained settings so that efficient optimization can be achieved.

\noindent \textbf{Keywords:} Bayesian optimization, Asynchronous optimization, Pipelining, Resource\textendash constrained experiment
\clearpage
\section{Introduction}

Parameter optimization methods search for parameters that yield results close to the objective by adjusting tunable parameters in a target system.
The criterion that indicates the objective is referred to as the objective function, and different parameter optimization methods are suitable for different properties of this function \cite{cauchy1847, Sibalija2019, katoch2021}.
When the objective function is explicitly expressed as a mathematical formula and the gradient of the function can be calculated, the steepest\textendash descent (or ascent) method is widely used, and it updates the solution in the direction of the gradient\cite{cauchy1847}.
In engineering and science, including biology and chemistry, the exact mathematical formula is often unknown, the gradient cannot be calculated, and obtaining experimental data can be costly \cite{wang2023}.
Particle swarm optimization \cite{Sibalija2019} and genetic algorithms \cite{katoch2021} are widely used optimization methods that do not use gradients.
These population\textendash based methods update a population of solutions and improve these solutions with each generation; however, they tend to require a large number of objective function evaluations, because a certain population size is required to avoid falling into a local optimum.
These methods are often used when the cost of obtaining the value of objective function is low, as in parameter optimization in computational simulations \cite{yang2014, sharafi2014, asgarpour_khansary2014}.

If the cost of obtaining experimental results is high, the number of evaluations needs to be reduced.
In such cases, response surface methodology (RSM) \cite{Bezerra2008} is used to model the objective function using polynomials to find the optimal parameters \cite{Ghafari2009,Fahimitabar2021}.
In RSM, experiments are carried out on the basis of experimental design such as central composite design \cite{box1951} or Box\textendash Behnken design \cite{box1960}.
These experimental designs uniformly sample from the parameter space, allowing for the creation of a model that can accurately predict the objective function across the entire parameter space.
However, RSM is time-consuming because experiments must be carried out even under conditions where the expectation of good results is low.

The Bayesian optimization method was developed to solve this issue \cite{kushner1964}.
In Bayesian optimization, the prior distribution of the objective function is combined with experimental results to calculate the posterior distribution for unexplored parameters.
Optimization is then performed by suggesting the next parameters by using an acquisition function that balances exploration and exploitation on the basis of the obtained posterior distribution.
Since Bayesian optimization proposes parameters on the basis of experimental results, it can densely explore the parameter space where there is a higher likelihood of achieving good results \cite{greenhill2020, liu2018}.
Unlike RSM, which explores the parameter space uniformly regardless of the results, Bayesian optimization can discover optimal parameters with fewer experimental evaluations \cite{Tachibana2023, claes_2024}.

\begin{figure}[h]
    \centering
    \includegraphics[scale=0.9]{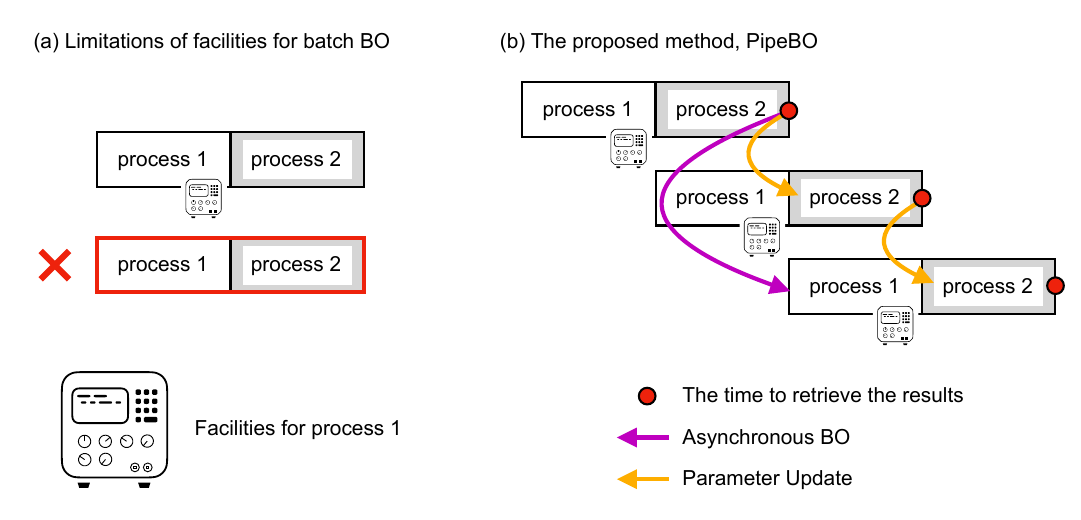}
    \caption{Conceptual diagram of parallelization through pipelining}
    \label{fig:overview}
    \raggedright{
        (a) Parameter optimization for an experiment with two sequential processes;
         a piece of equipment is required for the first process, but only one such piece is available.
        The processing time of optimization cannot be reduced using batch Bayesian optimization.
        (b) Pipelining to stagger the execution of processes allows experiments to be parallelized even with limited equipment.
        PipeBO aims to reduce the processing time of optimization.
        In applying Bayesian optimization, we focused on the ability to obtain a result from a completed experiment while another experiment is in progress.
        This led us to develop a method for parameter updates that leverages this ability.
        PipeBO combines the parameter updates with asynchronous Bayesian optimization
    }
\end{figure}

Vanilla Bayesian optimization is able to suggest only one experimental parameter, and it is necessary to iterate sequentially between obtaining a result and suggesting a parameter.
Batch Bayesian optimization \cite{ginsbourger2010, gonzalez2016, wu2016} focuses on reducing processing time of optimization by running parallel experiments with multiple parameters, but it is not efficient if the number of parallel experiments is limited due to resource constraints (Fig. \ref{fig:overview}a).
For example, there is a task to obtain a large quantity of useful enzymes, such as inulinase \cite{singh2018}.
The number of parallel experiments is limited by the number of flasks that fit into a rotary shaking incubator and by the number of incubators, so it is not possible to reduce processing time of optimization through the parallelization used by batch Bayesian optimization.

Many experiments involve multiple processes, with each process requiring different equipment. We found an analogy between this laboratory process and the pipelining used as a parallel processing method in central processing unit architecture (reviewed in \cite{ramamoorthy1977}).
We hypothesized that introducing the concept of pipelining into Bayesian optimization would reduce the processing time of optimization.
Pipelining enables the parallelization of experiments by starting processes while others are in progress (Fig. \ref{fig:overview}b) even when synchronous methods such as batch Bayesian optimization cannot be applied.

In this study, we proposed a Bayesian optimization method that reduced the processing time of optimization through parallelization via pipelining in settings with limited equipment.
We developed Pipelining Bayesian Optimization (PipeBO), that uses the results of other experiments that were obtained during a particular experiment (Fig. \ref{fig:overview}b).
To evaluate whether PipeBO can efficiently explore the parameter space, we used Black-Box Optimization Benchmarking (BBOB) \cite{hansen2009}.
We compared the processing time of optimization taken by PipeBO to reach conditions equivalent to those achieved by the vanilla methods.
PipeBO reduced the processing time of optimization to an average of $\leq$56\% for 20 out of the 24 functions, thereby demonstrating its efficacy.
\section{Methods}

\subsection{Problem setting for PipeBO}
An optimization problem searches for the parameter $\mathbf{x}^*$, that maximizes the experimental outcome (objective function) $f$ by adjusting the experimental parameters $\mathbf{x}$.
\begin{equation}
  \mathbf{x}^* = \underset{\mathbf{x}} {\operatorname{argmax}} \, f(\mathbf{x})
\end{equation}

Here, each experiment is divided into $K$ processes, with parameters that are set for each process (Supplementary Fig. \ref{fig:problem_setting}a).
We denote $\mathbf{s}_i\,(1\leq i \leq K)$ as a parameter set in the $i$-th process and we defined $\mathbf{s}_i$ as process parameters;
$\mathbf{x}$ are represented as $\mathbf{x} = \left(\mathbf{s}_1, \ldots, \mathbf{s}_K\right)$.
We denote the number of parameters for the $i$-th process as $N_i = \dim(\mathbf{s}_i)$, and we represent the sets of the numbers of process parameters for a single experiment as $D = (N_1, N_2, \ldots, N_K)$.
The result $ f(\mathbf{x}) $ can only be obtained upon completion of the $ K $-th process.

The improvement of efficiency owing to pipelining depends on the longest process, and the efficiency is maximized when each process takes the same time \cite{ramamoorthy1977};
we assumed that this time was equal in this problem setting.
We defined the unit time required for each process as 1 step, and the total elapsed time as the number of steps.

If equipment is limited and experiments cannot be parallelized, dividing experiments into $K$ processes allows for pseudo parallelization as a $K$-stage pipelining.
If $P$ experiments can be parallelized, pipelining can be incorporated, allowing for the parallelization of $P \times K$ experiments (Supplementary Fig. \ref{fig:problem_setting}b).
Among the $P \times K$ parallel experiments, we refer to the $P$ parallel experiments that have been executing the same process simultaneously as an experimental set, and we denote the $n$-th experimental set as $\mathbf{B}_n$.
The latter consists of $P$ parallel experimental parameters, with the $j$-th experimental parameter among them denoted as $\mathbf{x}_{j, n}\,(1\leq j \leq P)$.
In other words, we can express $\mathbf{B}_n$ as $\left(\mathbf{x}_{1,n}, \ldots, \mathbf{x}_{P,n}\right)$.
Then, if we denote the $i$-th process parameter that consists of $\mathbf{x}_{j, n}$ as $\mathbf{s}_{i_{j, n}}$, we can express $\mathbf{B}_n$ as $\left( \left(\mathbf{s}_{1_{1, n}}, \ldots,  \mathbf{s}_{K_{1, n}}\right), \ldots, \left(\mathbf{s}_{1_{P, n}}, \ldots,  \mathbf{s}_{K_{P, n}}\right) \right) $.

\subsection{PipeBO algorithm}
PipeBO is aimed to accelerate Bayesian optimization by using pipelining for the parallelization of experiments when equipment is insufficient and straightforward parallelization is not feasible.
Pipelining gives us results from an experiment while other experiments are in progress.
This property inspired us to develop a new method for updating parameters of an experiment in progress and to incorporate it into PipeBO.

Parallelization through pipelining is asynchronous: it allows the initiation of an experiment while another one is in progress.
In asynchronous experiments, it is necessary to propose the next experimental parameters when the results of some experiments have not yet been obtained.
Alvi {\itshape et al.} proposed asynchronous batch Bayesian optimization, which can efficiently explore the parameter space \cite{alvi2019}.
This method uses a local penalizer \cite{gonzalez2016} to prevent the proposal of parameters similar to those in the experiment in progress.
The local penalizer reduces the value of the acquisition function locally near certain experimental parameters.

The method proposed by Alvi {\itshape et al.} focuses on the parallelization of experiments where the required time depends considerably on the experimental parameters, but it does not divide a single experiment into multiple processes.
In contrast, in parallelization using PipeBO, the new results of asynchronously parallelized experiments are obtained during an experiment when subsequent process parameters can be changed.
To take advantage of these results, we developed a method to update the process parameters during the experiment.
Specifically, the acquisition function is recalculated on the basis of the data that include new results, and the process parameters are updated to give the best value of the new acquisition function.

\begin{figure}
    \centering
    \includegraphics[scale=0.8]{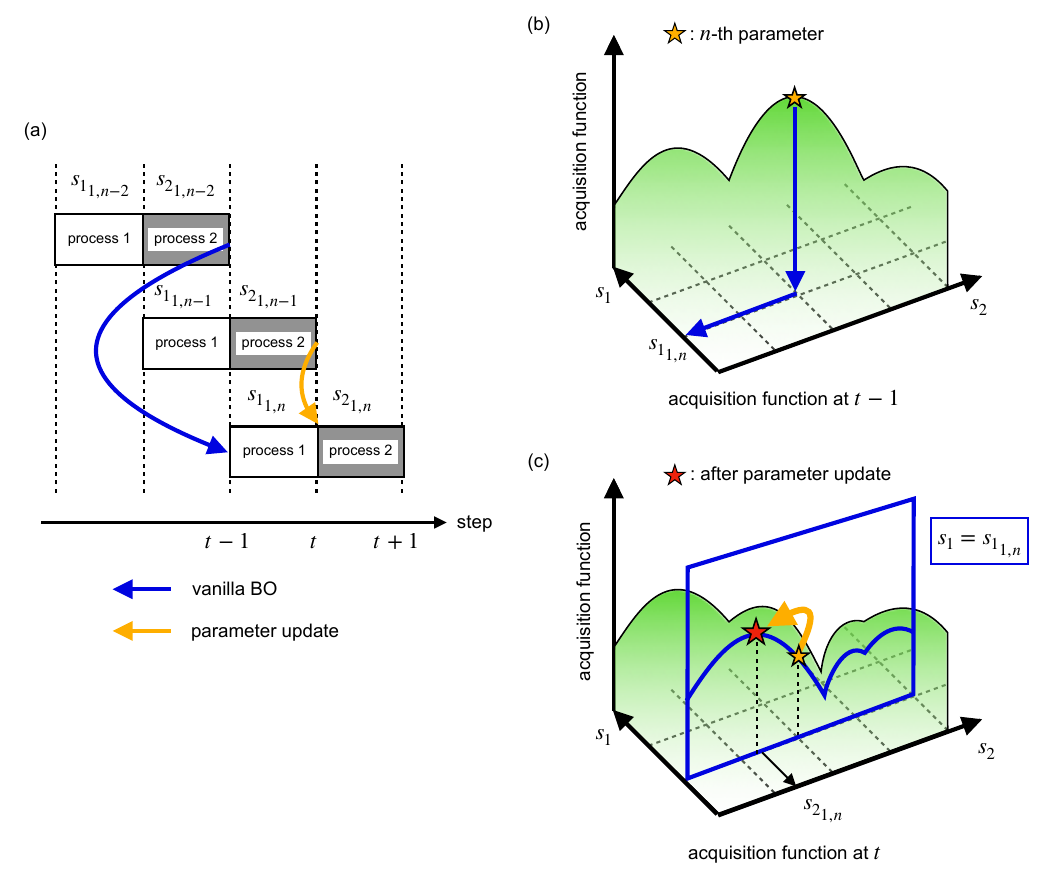}
    \caption{Updating parameters during the experiment}
    \label{fig:update}
    \raggedright{
        (a) Flow of Bayesian optimization incorporating pipelining and use of information at each step.
        At $t-1$ step, the experimental parameters are proposed using vanilla Bayesian optimization (blue arrow).
        In the $t$ step, pipelining allows for new experimental results to be obtained, which are then used to update the second process parameter of the $n$-th experiment (yellow arrow).
        (b) The acquisition function based on the results of up to the $(n-2)$-th experiment, using the results from the $t-1$ step (green space).
        Since the results from the $(n-1)$-th experiment are not yet available at the $t-1$ step, the computation of the acquisition function is based on the results from the $(n-2)$-th experiment to propose the experimental parameters.
        (c) The acquisition function based on the results of up to the $(n-1)$-th experiment (green space).
        The blue frame is the plane that is $s_1 = s_{1_{1,n}}$, the blue curve is the acquisition function cut by the plane.
        At the $t$ step, new results from the $(n-1)$-th experiment are obtained, causing the acquisition function to change from that shown in (b).
        Since the $n$-th experiment is already underway with the process parameter $s_{1_{1,n}}$, $s_{1_{1,n}}$ is fixed to improve the acquisition function value, and $s_{2_{1,n}}$ is updated (yellow arrow)
    }
\end{figure}

As a simple problem setting, consider that the parallelization number is $P=1$, the number of processes is $K=2$, and the process parameters $s_1$ for the first process and $s_2$ for the second can be adjusted.
In the $n$-th experiment (Fig. \ref{fig:update}a), the first process parameter $s_{1_{1,n}}$ must be determined after the results of the $(n-2)$-th experiment have been obtained.
Similar to vanilla Bayesian optimization, we determine $s_{1_{1,n}}$ by calculating the acquisition function based on the results obtained from up to the $(n-2)$-th experiment, and we propose the experimental parameter $\mathbf{x}_{1,n}$ (blue arrow in Fig. \ref{fig:update}a, b).
After this step, the first process of the $n$-th experiment is executed with $\mathbf{x}_{1,n}$.
Simultaneously, the result of the $(n-1)$-th experiment is obtained,
so a new acquisition function is derived on the basis of the results obtained up to the $(n-1)$-th experiment (Fig. \ref{fig:update}c).
In this recalculated acquisition function with new results, there may be a value greater than the value of $\mathbf{x}_{1,n}$ based on the acquisition function derived from the results up to the $(n-2)$-th experiment.
Therefore, during the $n$-th experiment, while keeping the already determined $s_{1_{1,n}}$ fixed, $s_{1_{2,n}}$ is adjusted to achieve a higher value of the acquisition function (yellow arrows in Fig. \ref{fig:update}a, c).
By updating the process parameters during experiments in this way, we have developed a method that optimizes an objective function while using the information obtained at each time point.

The pseudo code of PipeBO is shown in Algorithm \ref{algo:pipelining_batch_bayesian_optimization}.
Let the set of experimental parameters in progress be denoted as $\mathbf{X}_\text{LP}$, and let the function that locally decreases the value of the acquisition function at the experimental parameter $\mathbf{x}_\text{LP} \in \mathbf{X}_\text{LP}$ be represented as $\phi(\mathbf{x};\mathbf{x}_\text{LP})$.
, which is expressed as follows \cite{gonzalez2016}:
\begin{equation}
  \label{eq:local_Penalizer}
  \phi(\mathbf{x};\mathbf{x}_\text{LP}) = \frac{1}{2} \mathrm{erfc} \left(- \frac{1}{2\sigma^{2}(\mathbf{x}_\text{LP})} \left(\hat{L} \| \mathbf{x}_\text{LP} - \mathbf{x} \| - \hat{M} + \mu(\mathbf{x}_\text{LP}) \right) \right)
\end{equation}
Here, $ \text{erfc}(\cdot) $ denotes the complementary error function,
$ \mu(\mathbf{x}) $ represents the posterior mean of the Gaussian process regression, $ \sigma^{2}(\mathbf{x}) $ indicates the posterior variance of this regression,
$ \hat{L} $ denotes the maximum slope at $ \mu(\mathbf{x}) $, and $ \hat{M} $ represents the maximum value obtained from the previous experimental results.

\begin{algorithm}
  \caption[Pipelining Bayesian Optimization]{Pipelining Bayesian Optimization (PipeBO)}
  \label{algo:pipelining_batch_bayesian_optimization}
  \begin{algorithmic}[1]
      \STATE Set initial conditions $\mathbf{B}_1, \mathbf{B}_2, \cdots, \mathbf{B}_K$
      \FOR {$t = K,K+1,...$,}
      \STATE Obtain experimental results for $\mathbf{B}_{t-K+1}$,
      \STATE Gaussian process regression with $\mathbf{B}_1, \cdots, \mathbf{B}_{t-K+1}$ and Obtain acquisition function $\alpha_t(\mathbf{x})$
      \STATE Initialization of parameter sets for local penalizer $\mathbf{X}_\text{LP} = \varnothing$
          \FOR {$i=1,\cdots, K-1$}
          \STATE Update $\mathbf{B}_{t-K+1+i}$ while keeping $({\mathbf s}_{1},\ldots,{\mathbf s}_{K-i})$ fixed in $\alpha_t(\mathbf{x}) {\displaystyle \prod_{\mathbf{x}_\text{LP} \in \mathbf{X}_\text{LP}}} \phi(\mathbf{x};\mathbf{x}_\text{LP})$
          \STATE Add $\mathbf{B}_{t-K+1+i}$ to $\mathbf{X}_\text{LP}$
          \ENDFOR
      \STATE Set $\mathbf{B}_{t+1}$ by Batch BO using  $\alpha_t(\mathbf{x}) {\displaystyle \prod_{\mathbf{x}_\text{LP} \in \mathbf{X}_\text{LP}}} \phi(\mathbf{x};\mathbf{x}_\text{LP})$
      \ENDFOR
  \end{algorithmic}
\end{algorithm}

After $ t $ steps, the results of the initial experiments are obtained, and the acquisition function $ \alpha_t(\mathbf{x}) $ at the $ t $ step is determined from these results.
Then, for the experimental sets in progress $ \mathbf{B}_{t-K+1+i} (i=1,\cdots, K-1) $, the previously determined process parameters $ ({\mathbf s}_{1},\ldots,{\mathbf s}_{K-i}) $ are fixed, and the undetermined parameters $({\mathbf s}_{K-i+1},\ldots,{\mathbf s}_{K})$ are adjusted to maximize the value of $ \alpha_t(\mathbf{x}) \prod_{\mathbf{x}_\text{LP} \in \mathbf{X}_\text{LP}} \phi(\mathbf{x};\mathbf{x}_\text{LP}) $ using the newly updated acquisition function.
The state in which $\mathbf{X}_\text{LP}$ contains no experimental parameters is denoted as the empty set $\mathbf{X}_\text{LP} = \varnothing$, and ${\displaystyle \prod_{\mathbf{x}_\text{LP} \in \mathbf{X}_\text{LP}}} \phi(\mathbf{x};\mathbf{x}_\text{LP})=1$ when $\mathbf{X}_\text{LP} = \varnothing$.
The updated experimental set $\mathbf{B}_{t-K+1+i}$ is added to $\mathbf{X}_\text{LP}$ to avoid proposing experimental parameters close to those in progress.
This operation is performed for all experimental parameters that were in progress, and finally the experimental set $\mathbf{B}_{t+1}$ to be used in the next experiment is proposed through batch Bayesian optimization.

\subsection{Numerical experiments using benchmark functions}
We evaluated the performance of PipeBO  by optimizing 24 benchmark functions included in BBOB \cite{hansen2009}, and compared the optimization processes.
These continuous functions of various shapes are defined over the domain [-5,5]$^D$ for any dimension $D$.
The BBOB functions were implemented in COmparing Continuous Optimizers (COCO) \cite{hansen2021}.

As the initial parameter values, $P \times K$ parameters were set for the method parallelized through pipelining, and $P$ parameters for the method without pipelining parallelization, using random numbers generated by Mersenne Twister \cite{matsumoto1998}.
For each benchmark function, the search for optimal experimental parameters was carried out 50 times with different initial parameter values.
Algorithms were compared using simple regret, which is the difference between the optimal result $f(\mathbf{x}^{*})$ in the benchmark function and the maximum value found in $n$ searches:
\begin{equation}
  \text{simple regret} = f(\mathbf{x}^{*}) - \underset{i=1,\cdots,n}{\text{max}}f(\mathbf{x}_i)
  \label{eq:simple_regret}
\end{equation}
A small simple regret indicates proximity to the optimum.
Therefore, the goal of optimization was to minimize the simple regret.

The acquisition function used was GP-UCB \cite{srinivas2010}, which is calculated by determining the upper confidence bound on the basis of the posterior mean $\mu(\mathbf{x})$ and the standard deviation $\sigma(\mathbf{x})$ from Gaussian process regression.
\begin{equation}
  \alpha_{\text{GP-UCB}} = \mu (\mathbf{x}) + \kappa \sigma ({\mathbf x})
\end{equation}
where $\kappa$ is a parameter related to the optimization strategy; it was fixed at $2$.

The 24 benchmark functions were optimized for the problem setup shown in Supplementary Table \ref{tb:verification_param};
the maximum number of searches starting from the initial values was 200.

The performance of PipeBO was verified through the above procedure.
Implementation of all methods was based on GPyOpt \cite{gpyopt2016}, and the source code and data are available at \url{https://github.com/funalab/PipeBO}.
\section{Results}
\subsection{Evaluating the effect of pipelining}
To demonstrate the effectiveness of parallelization through pipelining in reducing the processing time of optimization, we compared PipeBO with vanilla Bayesian optimization.
For each method, we visualized the relationship between the number of steps and the simple regret by performing 50 runs of optimization from different initial values.
PipeBO tended to achieve lower simple regret faster than vanilla Bayesian optimization did (Fig. \ref{fig:result1}).
A similar trend was observed for all 24 benchmark functions in Supplementary Figs \ref{fig:process_all_2}\textendash \ref{fig:process_all_5}.

In pipelining, the number of experiments carried out in parallel increases in proportion to the number of processes, $ K $.
To evaluate whether the number of steps (i.e., elapsed time) required for optimization with PipeBO is reduced as $ K $ increases, we optimized benchmark functions by varying the $ K $ value.
With an increase in $ K $, the number of steps required for optimization with PipeBO was considerably reduced compared to that with vanilla Bayesian optimization (Fig. \ref{fig:result1}).
This result may be explained by the fact that PipeBO runs $P \times K$ experiments in parallel, so the larger $K$ is, the more experiments can be parallelized.
It was suggested that more parallelized experiments with pipelining could reduce the processing time of optimization.

We evaluated the number of steps PipeBO required to reach the same simple regret as the vanilla Bayesian optimization method reached after 100 steps.
PipeBO required fewer steps in almost all benchmark functions (Table \ref{tb:result_quantitative}).
The median number of steps could be calculated for 23 benchmark functions and was reduced by PipeBO to an average of about 56\% for $K=2$, about 50\% for $K=3$, and about 38\% for $K=5$ (Table \ref{tb:result_quantitative}).

\subsection{Evaluating the effect of parameter update}
\begin{figure}
    \centering
    \includegraphics[scale=1.4]{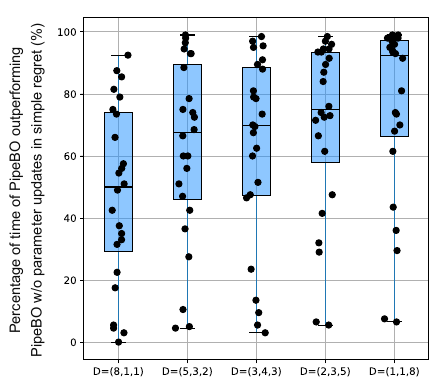}
    \caption{Comparison between PipeBO and PipeBO without parameter update}
    \label{fig:result3}
    \raggedright{
        To evaluate the effectiveness of parameter updates during an experiment, we determined PipeBO superiority on the basis of the median simple regret at each step and calculated the proportion of superiority in 200 iterations for each benchmark function.
        Each series indicates the number of parameters being optimized by each process.
        We used $D$ at $K = 3$ to compare cases where the optimized parameters were biased toward the earlier or later stages of the process.
        Datapoints in each series correspond to 24 benchmark functions, and the box plots represent the distribution of these data
    }
\end{figure}

To verify the effectiveness of updating parameters during the experiment in increasing the efficiency of optimization, we compared PipeBO and its version without parameter updates.
The parameter updates use the results from a completed experiment to update the parameters in a running experiment, and their performance may depend greatly on the sets of the numbers of process parameters.
Therefore, we used problem settings with biased sets of process parameters (Supplementary Table \ref{tb:verification_param}).
PipeBO without parameter update is equivalent to the asynchronous batch Bayesian optimization algorithm PLAyBOOK-L \cite{alvi2019}.
For PipeBO and PipeBO without parameter update, we determined their superiority on the basis of the median of simple regret at each step and calculated the ratio of superiority over 200 steps for each benchmark function (Fig. \ref{fig:result3}).
When the optimization parameters were biased toward the earlier processes, as in the case of $D=(8,1,1)$, the median was around 50\%, indicating little difference in performance of PipeBO with or without parameter update.
As the optimization parameters increasingly shifted toward the later processes, PipeBO had a higher percentage of steps where the simple regret was smaller than that of PipeBO without parameter update.
\section{Discussion}
Here we developed PipeBO, a Bayesian optimization method that incorporates pipelining to reduce the processing time of optimization when sufficient parallelization of experiments is impossible because experimental resources are limited.
Even under such conditions, PipeBO allows for effective parallelization (Fig. \ref{fig:overview}b).
To explore optimal parameters, PipeBO required fewer steps than vanilla Bayesian optimization without pipelining (Fig. \ref{fig:result1}) and required fewer steps to reach comparable simple regret (Table \ref{tb:result_quantitative}).

PipeBO reduced the number of optimization steps to about 56\% at $K=2$, about 50\% at $K=3$, and about 38\% at $K=5$.
In an ideal scenario where the optimization efficiency per experiment is equal between PipeBO and vanilla Bayesian optimization, the number of optimization steps would be reduced to $(100/K) \%$ (50\% at $K=2$, about 33\% at $K=3$, and 20\% at $K=5$)
 because PipeBO has $K$ times more experiments to parallelize.
However, to propose the n-th experimental set, vanilla Bayesian optimization uses the results of the $n-1$ experimental sets, whereas PipeBO uses only the results of up to the $n-K$ experimental sets.
This means that PipeBO has less information available to propose the parameters, and the number of steps required for optimization is not expected to be reduced to as much as  $(100/K) \%$.
The results of numerical experiments also supported this conclusion.

In lengthy experiments, the benefits of reducing the number of optimization steps are great.
For example, in enzymes production, fungal sources are cultured on potato dextrose agar (1st process) for 5 days, and then in flasks (2nd process) for 5 days \cite{singh2018}.
This is a $K=2$ pipeline process that can be parallelized.
If sequential Bayesian optimization takes 100 steps (500 days), PipeBO can reach same simple regret in 56 steps (280 days).

The pipelining process is similar to asynchronous batch Bayesian optimization \cite{alvi2019, kandasamy2018} in that another experiment is started during a particular experiment, but this study is unique in that (1) experiments are parallelized by pipelining and (2) the parameters are updated while an experiment is underway by using the results of other experiments.
Asynchronous batch Bayesian optimization was developed to avoid having equipment idle and reduce processing time of optimization for experiments where the required time depends considerably on the experimental parameters.
PipeBO differs from asynchronous batch Bayesian optimization in that PipeBO enables parallelization of experiments where experimental time is independent of parameters.
Vanilla Bayesian optimization methods and asynchronous batch Bayesian optimization typically assume that all parameters must be determined before initiating an experiment.
However, when an experiment can be divided into multiple processes, it is sufficient to determine the parameters at the beginning of each process.
During the execution of a particular process in pipelining, results from experiments with different parameters can be obtained simultaneously.
We focused on this point and developed a method to update parameters during an experiment within the pipelining framework (Fig. \ref{fig:update}).
In particular, when numerous parameters had to be optimized in later processes, the percentage of steps in which PipeBO achieved smaller simple regret increased (Fig. \ref{fig:result3}).
The reason might be that setting more parameters in the later stages increases the likelihood of finding good results within a wider parameter space.

PipeBO may not be suitable in experiments where the time required for each process varies considerably, because the benefits of parallelization through pipelining may be limited.
In such experiments, scheduling algorithms would be useful for reducing the processing time of optimization.
A scheduling algorithm reduces the time required to complete experimental procedures by optimizing the assignment of processes to the equipment \cite{itoh2021}.
Bayesian optimization based on schedules formulated by these algorithms may reduce the time even in problem settings where batch Bayesian optimization and PipeBO are not applicable.
Scheduling algorithms often parallelize experiments asynchronously so that the results of one experiment are often obtained during another experiment \cite{itoh2021, arai2023}.
Therefore, the parameter updates proposed in this study can be used in such schedules and can be applicable not only to pipelining but also to a broader range of methods.

This study will contribute to reducing the processing time of optimization in problem settings that reflect limited experimental facilities.
By helping to reduce the processing time of optimization, this study may play an important role in the advancement of data-driven science in fields where optimization has not been attempted due to the extensive time required for experiments.

\section*{Acknowledgement}
The research was funded by Japan Science and Technology Agency Core Research for Evolutionary Science and Technology (JST CREST), Japan (Grant Number JPMJCR21N1) to A.F.

\section*{Contributions}
\textbf{Y. T.:} Data curation, Formal analysis, Investigation, Methodology, Software, Validation, Visualization, Writing\textendash original draft, Writing\textendash review \& editing.
\textbf{Y. S.:} Formal analysis, Software, Visualization.
\textbf{Y. H.:} Formal analysis, Visualization, Writing\textendash review \& editing.
\textbf{T. M.:} Formal analysis, Visualization, Writing\textendash review \& editing.
\textbf{T. G. Y.:} Conceptualization, Formal analysis, Methodology, Supervision, Visualization, Writing\textendash review \& editing.
\textbf{A. F.:} Conceptualization, Formal analysis, Funding acquisition, Methodology, Project administration, Resources, Supervision, Visualization, Writing\textendash review \& editing.

\section*{Competing Interests}
The authors declare no competing interests.

\printbibliography

@article{Bezerra2008,
title = {{Response surface methodology (RSM) as a tool for optimization in analytical chemistry}},
journal = {Talanta},
volume = {76},
number = {5},
pages = {965-977},
year = {2008},
doi = {https://doi.org/10.1016/j.talanta.2008.05.019},
author = {Marcos Almeida Bezerra and Ricardo Erthal Santelli and Eliane Padua Oliveira and Leonardo Silveira Villar and Luciane Am\'{e}lia Escaleira},
}

@article{box1951,
	title = {On the {Experimental} {Attainment} of {Optimum} {Conditions}},
	volume = {13},
	number = {1},
	journal = {J. R. Stat. Soc. Ser. B},
	author = { George E. P. Box and Wilson, K. B.},
	year = {1951},
	pages = {1--38},
    doi = {https://doi.org/10.1111/j.2517-6161.1951.tb00067.x}
}

@article{box1960,
	title = {Some {New} {Three} {Level} {Designs} for the {Study} of {Quantitative} {Variables}},
	volume = {2},
	number = {4},
	journal = {Technometrics},
	author = {George E. P. Box and Behnken, D. W.},
	year = {1960},
	pages = {455--475},
    doi = {https://doi.org/10.2307/1266454}
}

@article{cauchy1847,
title = {M\'{e}thode g\'{e}n\'{e}rale pour la r\'{e}solution des syst\`{e}mes d'\'{e}quations simultan\'{e}es},
journal = {Comptes Rendus Hebd. S\'{e}ances Acad. Sci.},
volume = {25},
pages = {536--538},
year = {1847},
author = {Augustin Cauchy}
}

@article{Sibalija2019,
title = {Particle swarm optimisation in designing parameters of manufacturing processes: A review (2008-2018)},
journal = {Appl. Soft Comput.},
volume = {84},
pages = {105743},
year = {2019},
doi = {https://doi.org/10.1016/j.asoc.2019.105743},
author = {Tatjana V. Sibalija},
}

@article{katoch2021,
	title = {A review on genetic algorithm: past, present, and future},
	volume = {80},
	shorttitle = {A review on genetic algorithm},
	doi = {https://doi.org/10.1007/s11042-020-10139-6},
	number = {5},
	journal = {Multimed. Tools and Applica.},
	author = {Katoch, Sourabh and Chauhan, Sumit Singh and Kumar, Vijay},
	year = {2021},
	pages = {8091--8126},
}

@article{yang2014,
    title = {An improved {PSO}-based charging strategy of electric vehicles in electrical distribution grid},
    journal = {Appl. Energy},
    volume = {128},
    pages = {82-92},
    year = {2014},
    doi = {https://doi.org/10.1016/j.apenergy.2014.04.047},
    author = {Jun Yang and Lifu He and Siyao Fu},
    }

@article{sharafi2014,
    title = {Multi-objective optimal design of hybrid renewable energy systems using {PSO}-simulation based approach},
    journal = {Renew. Energy},
    volume = {68},
    pages = {67-79},
    year = {2014},
    doi = {https://doi.org/10.1016/j.renene.2014.01.011},
    author = {Masoud Sharafi and Tarek Y. ELMekkawy},
    }

@article{asgarpour_khansary2014,
    title = {Using genetic algorithm ({GA}) and particle swarm optimization ({PSO}) methods for determination of interaction parameters in multicomponent systems of liquid–liquid equilibria},
    journal = {Fluid Phase Equilib.},
    volume = {365},
    pages = {141-145},
    year = {2014},
    doi = {https://doi.org/10.1016/j.fluid.2014.01.016},
    author = {Milad Asgarpour Khansary and Ahmad Hallaji Sani},
    }

@article{kushner1964,
    author={Kushner, Harold J.},
    title={{A New Method of Locating the Maximum Point of an Arbitrary Multipeak Curve in the Presence of Noise}},
    journal={J. Basic Eng.},
    volume = {86},
    number = {1},
    pages = {97-106},
    year = {1964},
    doi = {https://doi.org/10.1115/1.3653121}
}

@Inbook{ginsbourger2010,
    author="Ginsbourger, David and Le Riche, Rodolphe and Carraro, Laurent",
    title={{Kriging Is Well-Suited to Parallelize Optimization}},
    bookTitle="Computational Intelligence in Expensive Optimization Problems",
    year="2010",
    publisher="Springer Berlin Heidelberg",
    volume = "131--162",
    doi = {https://doi.org/10.1007/978-3-642-10701-6_6}
}

@InProceedings{gonzalez2016,
    title = 	 {{Batch Bayesian Optimization via Local Penalization}},
    author = 	 {Gonz\'{a}lez, Javier and Dai, Zhenwen and Hennig, Philipp and Lawrence, Neil},
    booktitle = 	 {Proceedings of the 19th International Conference on Artificial Intelligence and Statistics},
    pages = 	 {648--657},
    year = 	 {2016},
    volume = 	 {51},
    series = 	 {Proceedings of Machine Learning Research},
    publisher =    {PMLR},
}

@inproceedings{wu2016,
    author = {Wu, Jian and Frazier, Peter},
    booktitle = {Advances in Neural Information Processing Systems},
    pages = {3126--3134},
    publisher = {Curran Associates, Inc.},
    title = {{The Parallel Knowledge Gradient Method for Batch Bayesian Optimization}},
    volume = {29},
    year = {2016}
}

@article{wang2023,
    author = {Wang, Xilu and Jin, Yaochu and Schmitt, Sebastian and Olhofer, Markus},
    title = {{Recent Advances in Bayesian Optimization}},
    year = {2023},
    issue_date = {December 2023},
    publisher = {Association for Computing Machinery},
    volume = {55},
    number = {13s},
    doi = {https://doi.org/10.1145/3582078},
    journal = {ACM Comput. Surv.},
    articleno = {287},
    numpages = {36}
}

@inproceedings{srinivas2010,
author = {Srinivas, Niranjan and Krause, Andreas and Kakade, Sham and Seeger, Matthias},
title = {Gaussian process optimization in the bandit setting: no regret and experimental design},
year = {2010},
publisher = {Omnipress},
booktitle = {Proceedings of the 27th International Conference on International Conference on Machine Learning},
volume = {1015--1022},
numpages = {8},
}

@article{greenhill2020,
    title = {Bayesian {Optimization} for {Adaptive} {Experimental} {Design}: {A} {Review}},
    volume = {8},
    shorttitle = {Bayesian {Optimization} for {Adaptive} {Experimental} {Design}},
    doi = {https://doi.org/10.1109/ACCESS.2020.2966228},
    journal = {IEEE Access},
    author = {Greenhill, Stewart and Rana, Santu and Gupta, Sunil and Vellanki, Pratibha and Venkatesh, Svetha},
    year = {2020},
    pages = {13937--13948},
    }

@article{liu2018,
    title = {A survey of adaptive sampling for global metamodeling in support of simulation-based complex engineering design},
    volume = {57},
    doi = {https://doi.org/10.1007/s00158-017-1739-8},
    number = {1},
    journal = {Struct. Multidiscip. Optim.},
    author = {Liu, Haitao and Ong, Yew-Soon and Cai, Jianfei},
    year = {2018},
    pages = {393--416},
    }

@article{Tachibana2023,
    author = {Tachibana, Ryo and Zhang, Kailin and Zou, Zhi and Burgener, Simon and Ward, Thomas R.},
    title = {{A Customized Bayesian Algorithm to Optimize Enzyme-Catalyzed Reactions}},
    journal = {ACS Sustain. Chem. Eng.},
    volume = {11},
    number = {33},
    pages = {12336-12344},
    year = {2023},
    doi = {https://doi.org/10.1021/acssuschemeng.3c02402},
}

@article{claes_2024,
	title = {Bayesian cell therapy process optimization},
	volume = {121},
	copyright = {© 2024 Wiley Periodicals LLC.},
	number = {5},
	journal = {Biotechnol. Bioeng.},
	author = {Claes, Evan and Heck, Tommy and Coddens, Kathleen and Sonnaert, Maarten and Schrooten, Jan and Verwaeren, Jan},
	year = {2024},
	pages = {1569--1582},
    doi = {https://doi.org/10.1002/bit.28669}
}

@InProceedings{alvi2019,
    title = 	 {{Asynchronous Batch Bayesian Optimisation with Improved Local Penalisation}},
    author =       {Alvi, Ahsan S. and Ru, Binxin and Calliess, Jan Peter and Roberts, Stephen J. and Osborne, Michael A.},
    booktitle = 	 {Proceedings of the 36th International Conference on Machine Learning},
    pages = 	 {253--262},
    year = 	 {2019},
    volume = 	 {97},
    series = 	 {Proceedings of Machine Learning Research},
    publisher =    {PMLR},
}

@InProceedings{kandasamy2018,
    title = 	 {{Parallelised Bayesian Optimisation via Thompson Sampling}},
    author = 	 {Kandasamy, Kirthevasan and Krishnamurthy, Akshay and Schneider, Jeff and Poczos, Barnabas},
    booktitle = 	 {Proceedings of the 21st International Conference on Artificial Intelligence and Statistics},
    pages = 	 {133--142},
    year = 	 {2018},
    volume = 	 {84},
    series = 	 {Proceedings of Machine Learning Research},
    publisher =    {PMLR},
}

@article{Ghafari2009,
title = {{Application of response surface methodology (RSM) to optimize coagulation-flocculation treatment of leachate using poly-aluminum chloride (PAC) and alum}},
journal = {J. Hazard. Mater.},
volume = {163},
number = {2},
pages = {650-656},
year = {2009},
doi = {https://doi.org/10.1016/j.jhazmat.2008.07.090},
author = {Shahin Ghafari and Hamidi Abdul Aziz and Mohamed Hasnain Isa and Ali Akbar Zinatizadeh},
}

@article{Fahimitabar2021,
title = {{Application of RSM for optimization of glutamic acid production by {\it Corynebacterium glutamicum} in bath culture}},
journal = {Heliyon},
volume = {7},
number = {6},
pages = {e07359},
year = {2021},
doi = {https://doi.org/10.1016/j.heliyon.2021.e07359},
author = {Azadeh Fahimitabar and Seyyed Mohammad Hossein Razavian and Seyyed Ali Rezaei},
}

@article{ramamoorthy1977,
    author = {Ramamoorthy, C. V. and Li, H. F.},
    title = {{Pipeline Architecture}},
    year = {1977},
    issue_date = {March 1977},
    publisher = {Association for Computing Machinery},
    address = {New York, NY, USA},
    volume = {9},
    number = {1},
    journal = {ACM Comput. Surv.},
    pages = {61--102},
    numpages = {42},
    doi = {https://doi.org/10.1145/356683.356687}
}

@article{singh2018,
    author = "Ram Sarup Singh and Kanika Chauhan",
    title = "Sequential statistical optimization of lactose–based medium and process variables for inulinase production from {\it Penicillium oxalicum} {BGPUP-4}",
    journal = "3 Biotech",
    year = "2018",
    volume = "8",
    number = "38",
    doi = "https://doi.org/10.1007/s13205-017-1060-7"
}

@Misc{gpyopt2016,
    author = {{The GPyOpt authors}},
    title = {{GPyOpt}: A Bayesian Optimization framework in python},
    year = {2016},
    howpublished = {\url{http://github.com/SheffieldML/GPyOpt}},
}

@article{hansen2009,
    title={{Real-Parameter Black-Box Optimization Benchmarking 2009: Noiseless Functions Definitions}},
    author={Nikolaus Hansen and Steffen Finck and Raymond Ros and Anne Auger},
    journal = {\rm {[Research Report] RR-6829, INRIA. inria-00362633v2, 16p.}},
    year = {2009}
}

@article{hansen2021,
    author = {Hansen, N. and Auger, A. and Ros, R. and Mersmann, O. and Tu{\v s}ar, T. and Brockhoff, D.},
    title = {{COCO}: A Platform for Comparing Continuous Optimizers in a Black-Box Setting},
    journal = {Optim. Methods and Softw.},
    doi = {https://doi.org/10.1080/10556788.2020.1808977},
    pages = {114--144},
    issue = {1},
    volume = {36},
    year = 2021
}

@article{itoh2021,
	title = {Optimal {Scheduling} for {Laboratory} {Automation} of {Life} {Science} {Experiments} with {Time} {Constraints}},
	volume = {26},
	number = {6},
	journal = {SLAS Technol.},
	author = {Itoh, Takeshi D. and Horinouchi, Takaaki and Uchida, Hiroki and Takahashi, Koichi and Ozaki, Haruka},
	year = {2021},
	pages = {650--659},
    doi = {https://doi.org/10.1177/24726303211021790}
}

@article{arai2023,
	title = {{SAGAS}: {Simulated} annealing and greedy algorithm scheduler for laboratory automation},
	volume = {28},
	shorttitle = {{SAGAS}},
	number = {4},
	journal = {SLAS Technol.},
	author = {Arai, Yuya and Takahashi, Ko and Horinouchi, Takaaki and Takahashi, Koichi and Ozaki, Haruka},
	year = {2023},
	pages = {264--277},
    doi = {https://doi.org/10.1016/j.slast.2023.03.001}
}

@article{matsumoto1998,
	title = {Mersenne twister: a 623-dimensionally equidistributed uniform pseudo-random number generator},
	volume = {8},
	shorttitle = {Mersenne twister},
	number = {1},
	journal = {ACM Trans. Model. Comput. Simul.},
	author = {Matsumoto, Makoto and Nishimura, Takuji},
	year = {1998},
	pages = {3--30},
    doi = {https://doi.org/10.1145/272991.272995}
}


\clearpage

\clearpage

\clearpage
\begin{figure}
    \centering
    \includegraphics[scale=0.9]{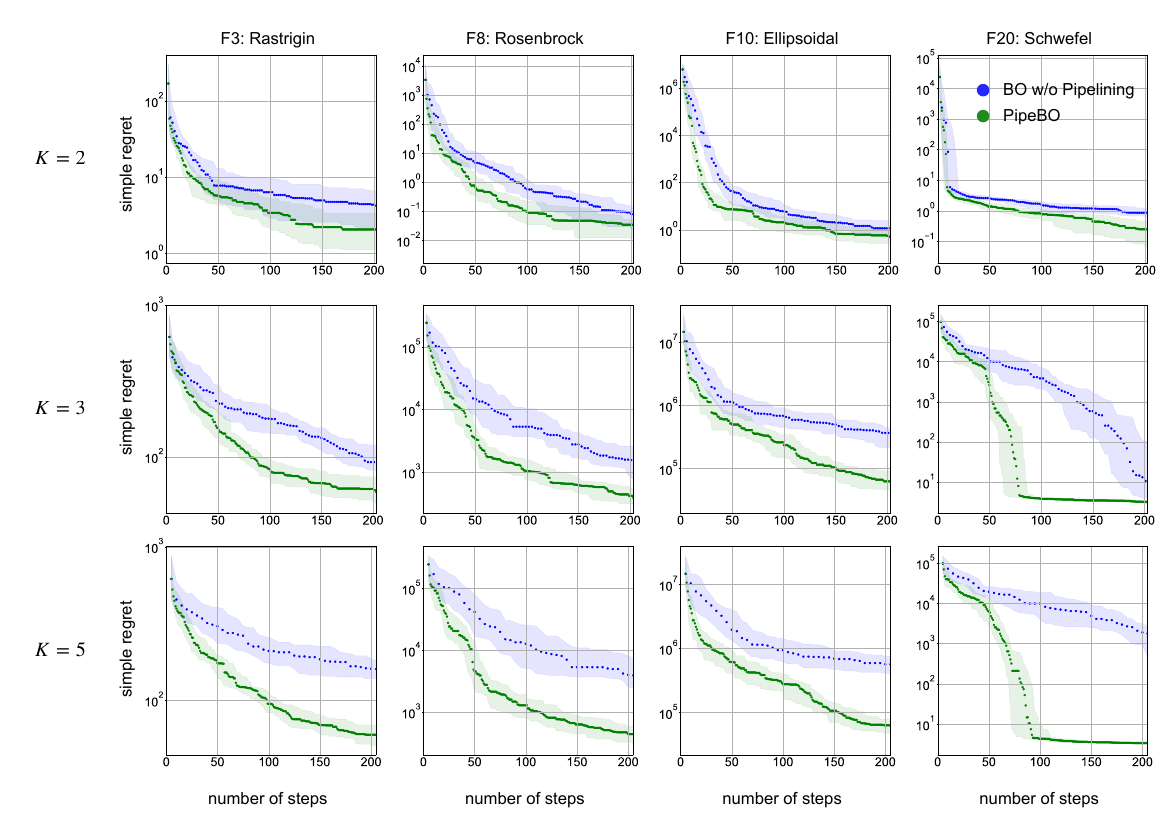}
    \caption{Comparison between Bayesian optimization (BO) without pipelining and PipeBO}
    \label{fig:result1}
    \raggedright{
        The median (lines) and interquartile range (shading) of 50 runs of optimization from different initial values are shown for each method.
        $K$, the number of processes; function IDs (F) are shown for the 4 functions selected from the 24 BBOB functions.
        All results are for $P$ = 1; $D = (1,1)$ at $K = 2$, $D = (3,4,3)$ at $K = 3$, and $D = (2,2,2,2,2)$ at $K = 5$
    }
\end{figure}

\clearpage
\begin{table}
    \begin{center}
        \caption{Quantitative evaluation of the number of steps by Bayesian optimization without pipelining and PipeBO}
        \label{tb:result_quantitative}
        \begin{tabular}{llD{(}{\,(}{5}D{(}{\,(}{5}D{(}{\,(}{5}}
            \hline
            \multicolumn{1}{l}{ID} &
            \multicolumn{1}{l}{Benchmark function} &
            \multicolumn{1}{c}{$K=2$} &
            \multicolumn{1}{c}{$K=3$} &
            \multicolumn{1}{c}{$K=5$} \\
            \hline
            F1  & Sphere                                          & 126($\textendash$) & \bm{40.0}(5) & \bm{39.0}(7) \\
            F2  & Separable Ellipsoidal                           & \bm{50.0}(35) & \bm{41.5}(55) & \bm{23.0}(64) \\
            F3  & Rastrigin, original                             & \bm{42.0}(59) & \bm{45.0}(26) & \bm{33.5}(23) \\
            F4  & Büche\textendash Rastrigin                      & \bm{44.5}(45) & 101(116) & \bm{87.0}(139) \\
            F5  & Linear Slope                                    & $\textendash$ & $\textendash$ & $\textendash$ \\
            F6  & Attractive Sector                         & \bm{71.0}($\textendash$) & \bm{38.5}(24) & \bm{32.0}(13) \\
            F7  & Step Ellipsoidal                                & \bm{64.5}(86) & \bm{34.5}(23) & \bm{34.0}(35) \\
            F8  & Rosenbrock, original                            & \bm{51.5}(27) & \bm{45.5}(15) & \bm{44.0}(21) \\
            F9  & Rosenbrock, rotated                             & \bm{44.0}(31) & \bm{68.0}(50) & \bm{39.5}(36) \\
            F10 & Ellipsoidal                                     & \bm{68.5}(52) & \bm{37.0}(38) & \bm{32.0}(32) \\
            F11 & Discus                                          & \bm{56.0}(48) & \bm{50.5}(62) & \bm{30.0}(55) \\
            F12 & Bent Cigar                                      & \bm{43.5}(51) & 103($\textendash$) & \bm{49.5}(159) \\
            F13 & Sharp Ridge                                     & \bm{39.5}(28) & \bm{40.5}(10) & \bm{33.0}(9) \\
            F14 & Different Powers                                & \bm{44.5}(15) & \bm{39.5}(26) & \bm{36.0}(25) \\
            F15 & Rastrigin, rotated                              & \bm{40.5}(29) & \bm{47.5}(36) & \bm{38.5}(19) \\
            F16 & Weierstrass                                     & \bm{67.0}(87) & \bm{42.0}(46) & \bm{26.0}(23) \\
            F17 & Schaffer's F7                                   & \bm{66.0}(74) & \bm{43.0}(35) & \bm{20.5}(17) \\
            F18 & Schaffer's F7, moderately ill\textendash conditioned       & \bm{59.5}(41) & \bm{42.0}(48) & \bm{30.0}(27) \\
            F19 & Composite Griewank\textendash Rosenbrock function F8F2     & \bm{38.5}(30) & \bm{37.0}(50) & \bm{61.0}(53) \\
            F20 & Schwefel                                        & \bm{45.5}(36) & \bm{48.0}(14) & \bm{45.0}(27) \\
            F21 & Gallagher's Gaussian 101\textendash me Peaks               & \bm{50.0}(59) & \bm{37.0}(46) & \bm{38.0}(16) \\
            F22 & Gallagher's Gaussian 21\textendash hi Peaks                & \bm{53.5}(49) & \bm{66.5}(104) & \bm{36.5}(21) \\
            F23 & Katsuura                                        & \bm{56.0}(86) & \bm{49.0}(66) & \bm{19.5}(25) \\
            F24 & Lunacek bi\textendash Rastrigin                            & \bm{70.5}(68) & \bm{41.0}(14) & \bm{38.0}(16) \\
            \hline \hline
            \multicolumn{1}{c}{} &
            \multicolumn{1}{c}{Avarage} &
            \multicolumn{1}{c}{56.2} &
            \multicolumn{1}{c}{49.5} &
            \multicolumn{1}{c}{37.6} \\
            \hline \hline
        \end{tabular}

    \end{center}
    \raggedright{
        The number of steps PipeBO required to reach the same simple regret as Bayesian optimization without pipelining achieved in 100 steps.
        The values are the median numbers of steps taken by PipeBO to reach the reference simple regret; the interquartile ranges are shown in parentheses.
        Values less than 100 steps are displayed in bold and values greater than 100 steps are displayed in plain text.
        The calculations were based on data from 200 searches.
        Dashes (\textendash) indicate that the reference simple regret could not be reached
    }
\end{table}

\clearpage

\clearpage
\setcounter{figure}{0}
\setcounter{table}{0}
\captionsetup[figure]{name=Supplementary Fig.}  
\captionsetup[table]{name=Supplementary Table}    

\clearpage
\begin{figure}[h!]
    \centering
    \includegraphics[scale=0.8]{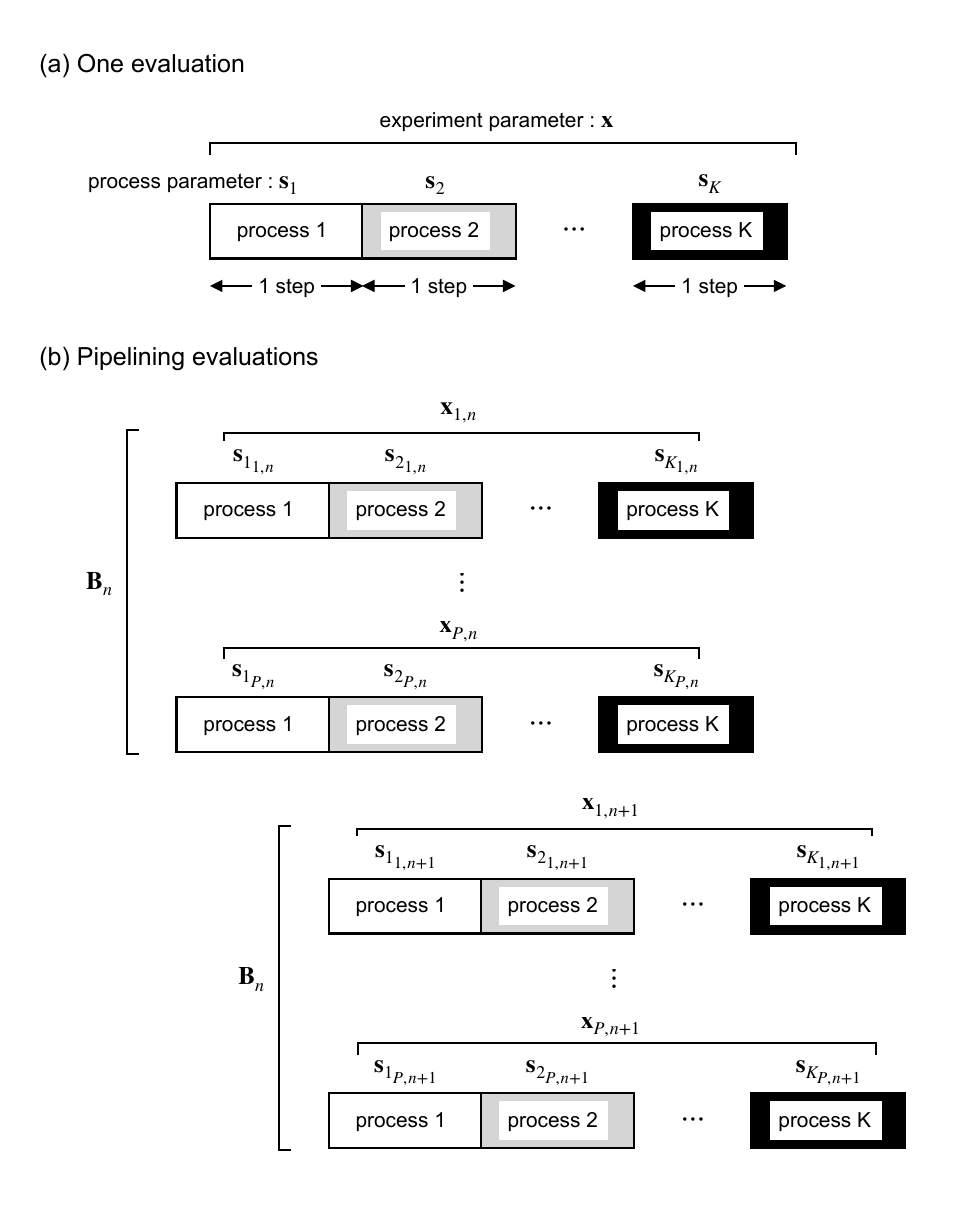}
    \caption{Problem sets tested in this study}
    \label{fig:problem_setting}
    \raggedright{
        (a) Parameter $\mathbf{x}$ of an experiment needs to be optimized.
        One experiment is divided into $K$ processes, and $\mathbf{x}$ is divided into $\mathbf{s}_i\,(1\leq i \leq K)$ parameters for the $i$-th process.
         ,where $\mathbf{x}$ is an experimental parameter and $\mathbf{s}_i$ is a process parameter.
        (b)Problem setting for experiments along the pipelining.
        The $P$ experiments that can be performed in parallel are called the experimental set $\mathbf{B}$, and the $n$-th experimental set $\mathbf{B}_n$ is divided into $P$ experimental parameters $\mathbf{x}_{j, n}\,(1\leq j \leq P)$.
        The $i$-th process parameter that makes up $\mathbf{x}_{j, n}$ is denoted as $\mathbf{s}_{i_{j, n}}$.
    }
\end{figure}

\clearpage
\begin{table}[h!]
    \begin{center}
        \caption{Problem sets to test the effectiveness of PipeBO}
        \label{tb:verification_param}
        \begin{tabular}[hb]{lll}
            \hline
            $P$ & $K$ & $D$ \\
            \hline
            1 & 2 & (1,\,1) \\
            1 & 3 & (8,\,1,\,1) \\
            1 & 3 & (5,\,3,\,2) \\
            1 & 3 & (3,\,4,\,3) \\
            1 & 3 & (2,\,3,\,5) \\
            1 & 3 & (1,\,1,\,8) \\
            1 & 5 & (2,\,2,\,2,\,2,\,2) \\
            \hline
        \end{tabular}
    \end{center}
    \raggedright{
        Since the number of parallel experiments increases in proportion to the number of processes $K$ in an experiment, we varied the $K$ value.
        We also varied $D$ because the parameter update is greatly affected by the set of the numbers of process parameters.
    }
\end{table}

\setlength{\topmargin}{-3.5cm}
\setlength{\footskip}{7cm}
\vspace*{-50cm}
\begin{figure}[h!]
    \centering
    \includegraphics[scale=0.96]{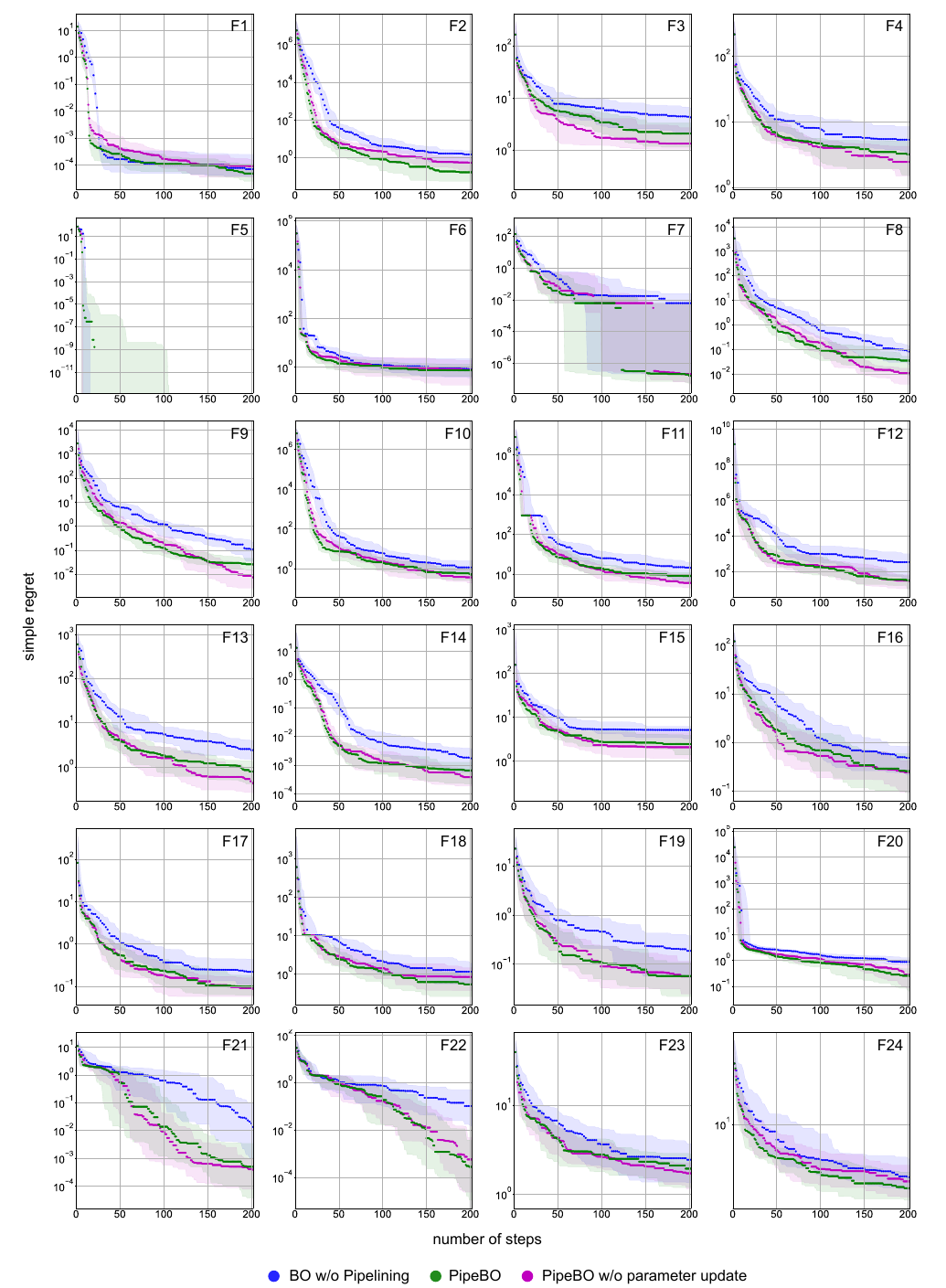}
    \caption{Optimization process for 24 benchmark functions ($K=2$)}
    \label{fig:process_all_2}
    \raggedright{
        The median (lines) and interquartile range (shading) of 50 runs of optimization from different initial values are shown for each function.
        The problem setting here was $P=1$, $K=2$ and $D=(1, 1)$.
        Function ID is shown in the upper right corner of each panel.
    }
\end{figure}

\vspace*{-50cm}
\begin{figure}[h!]
    \centering
    \includegraphics[scale=0.96]{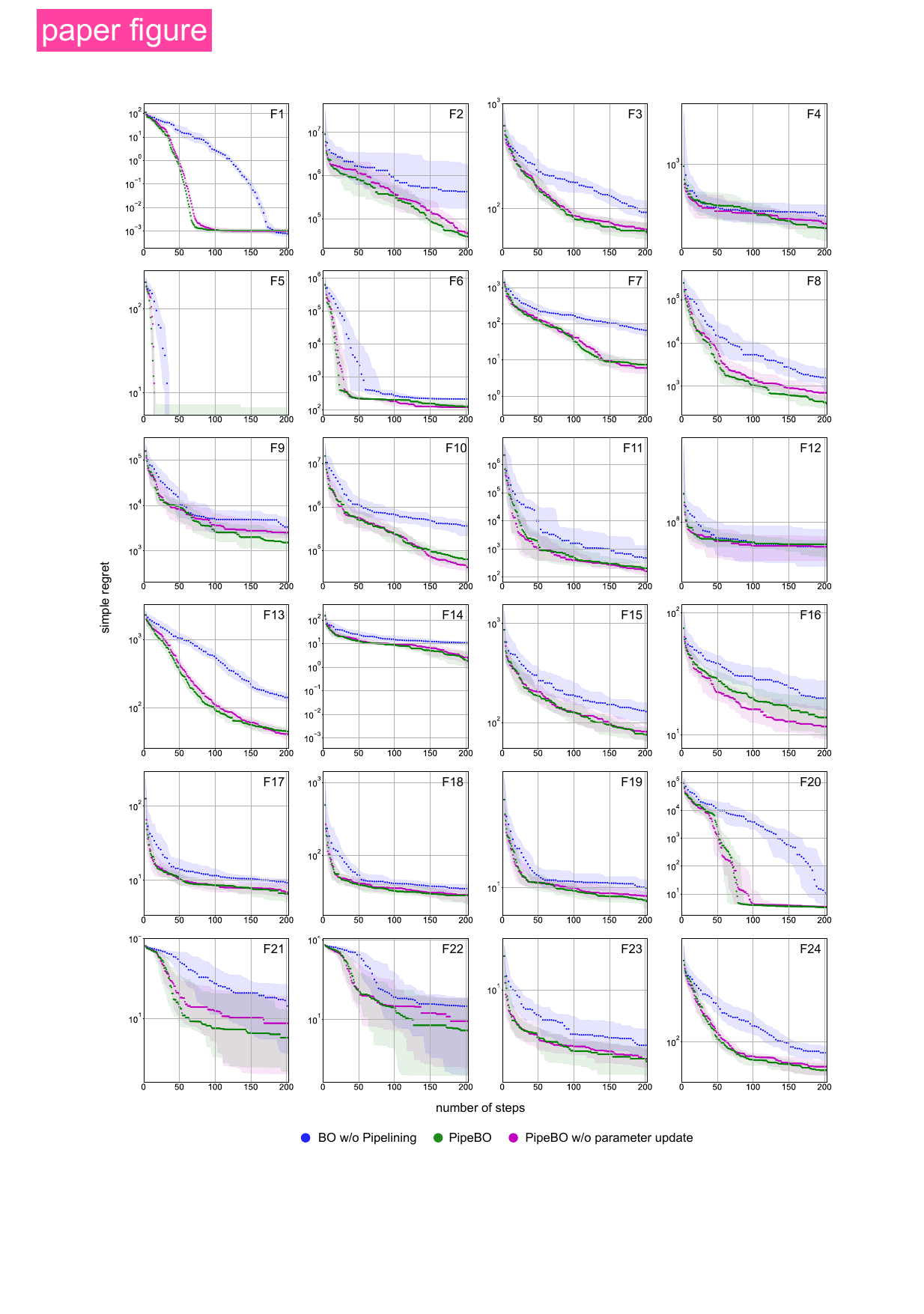}
    \caption{Optimization process for 24 benchmark functions ($K=3$)}
    \label{fig:process_all_3}
    \raggedright{
        The median (lines) and interquartile range (shading) of 50 runs of optimization from different initial values are shown for each function.
        The problem setting here was $P=1$, $K=3$ and $D=(3, 4, 3)$.
        Function ID is shown in the upper right corner of each panel.
    }
\end{figure}

\vspace*{-50cm}
\begin{figure}[h!]
    \centering
    \includegraphics[scale=0.96]{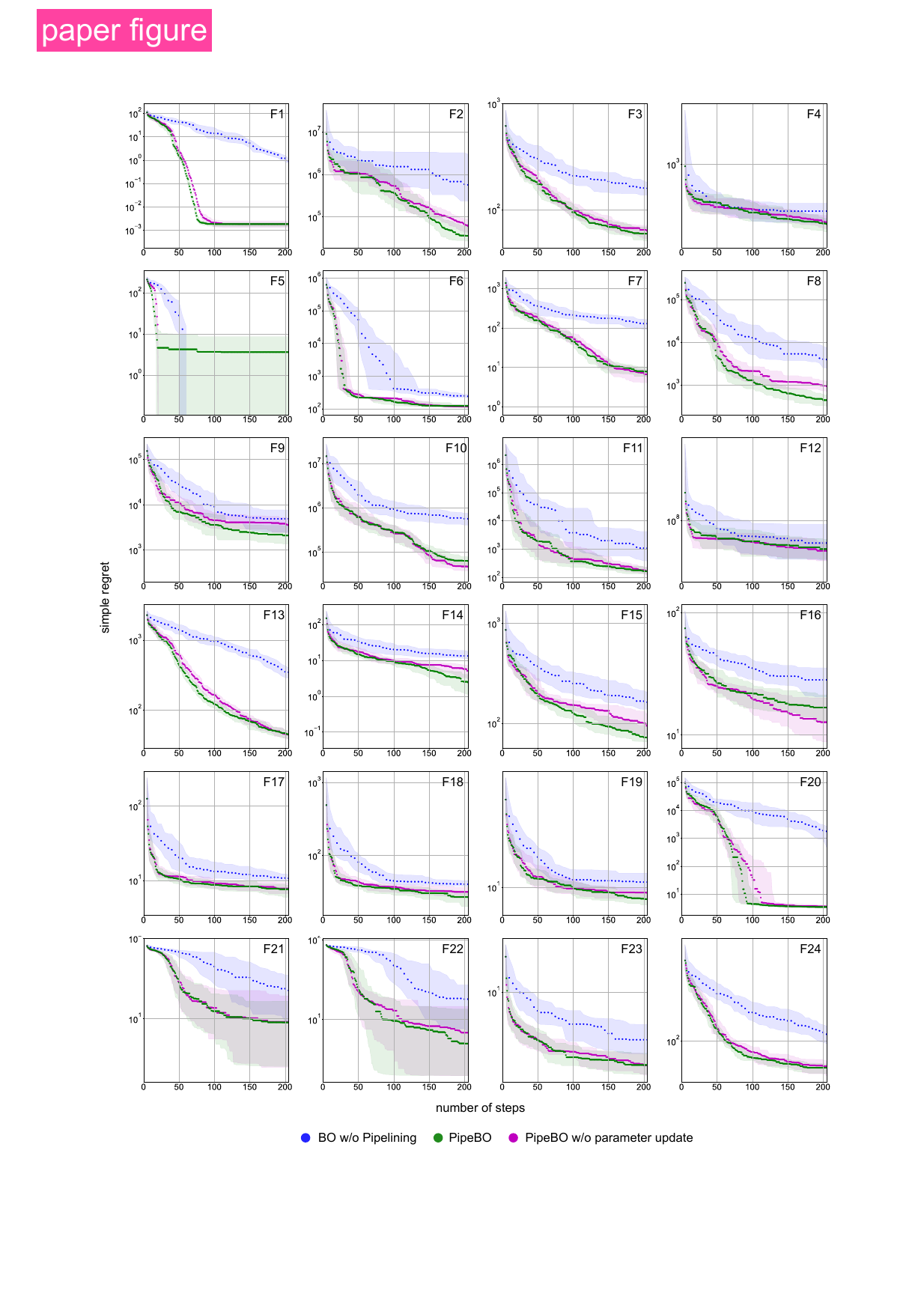}
    \caption{Optimization process for 24 benchmark functions ($K=5$)}
    \label{fig:process_all_5}
    \raggedright{
        The median (lines) and interquartile range (shading) of 50 runs of optimization from different initial values are shown for each function.
        The problem setting here was $P=1$, $K=5$ and $D=(2, 2, 2, 2, 2)$.
        Function ID is shown in the upper right corner of each panel.
    }
\end{figure}
\end{document}